\documentclass[10pt,twocolumn,letterpaper]{article}

\usepackage[pagenumbers]{cvpr} 
\usepackage{array} 
\usepackage{multicol} 
\usepackage{multirow} 
\usepackage{makecell}
\usepackage{bbding}
\usepackage{bbm}
\usepackage{amssymb}
\usepackage{colortbl}
\usepackage[table]{xcolor} 
\usepackage{xcolor}
\definecolor{SpringGreen4}{RGB}{46,139,87}
\definecolor{Honeydew4}{RGB}{131, 139, 131} 
\definecolor{SeaGreen4}{RGB}{0,205,102} 
\definecolor{SlateBlue}{RGB}{106,90,205} 
\definecolor{DarkRed}{RGB}{178,34,34} 

\definecolor{cvprblue}{rgb}{0.21,0.49,0.74}
\usepackage[pagebackref,breaklinks,colorlinks,allcolors=cvprblue]{hyperref}

\def\paperID{*****} 
\def\confName{CVPR}
\def\confYear{2025}

\title{
UniPAR: A Unified Framework for Pedestrian Attribute Recognition 
}

\author{Minghe Xu$^{1,2}$, Rouying Wu$^{3}$, Jiarui Xu$^{4}$, Minhao Sun$^{4}$, Zikang Yan$^{4}$, \\ 
    Xiao Wang$^{4}$, ChiaWei Chu$^{1}$, Yu Li$^{2}$\thanks{Corresponding Author: Yu Li} \\ 
${^1}$ Faculty of Data Science, City University of Macau, Macau SAR, China \\ 
${^2}$ School of Big Data, Zhuhai College of Science and Technology, Zhuhai 519041, China \\ 
${^3}$ Faculty of Innovation Engineering, Macau University of Science and Technology, Macau SAR, China \\ 
${^4}$ School of Computer Science and Technology, Anhui University, Hefei 230601, China \\ 
\textit{xuminghe001@foxmail.com}
} 

\begin{document}
\maketitle

\begin{abstract}
Pedestrian Attribute Recognition is a foundational computer vision task that provides essential support for downstream applications, including person retrieval in video surveillance and intelligent retail analytics. However, existing research is frequently constrained by the ``one-model-per-dataset" paradigm and struggles to handle significant discrepancies across domains in terms of modalities, attribute definitions, and environmental scenarios. To address these challenges, we propose UniPAR, a unified Transformer-based framework for PAR. By incorporating a unified data scheduling strategy and a dynamic classification head, UniPAR enables a single model to simultaneously process diverse datasets from heterogeneous modalities, including RGB images, video sequences, and event streams. We also introduce an innovative phased fusion encoder that explicitly aligns visual features with textual attribute queries through a late deep fusion strategy. Experimental results on the widely used benchmark datasets, including MSP60K, DukeMTMC, and EventPAR, demonstrate that UniPAR achieves performance comparable to specialized SOTA methods. Furthermore, multi-dataset joint training significantly enhances the model's cross-domain generalization and recognition robustness in extreme environments characterized by low light and motion blur. The source code of this paper will be released on \url{https://github.com/Event-AHU/OpenPAR}
\end{abstract}

\section{Introduction} 
Pedestrian Attribute Recognition (PAR)~\cite{wang2021pedestrian} is a foundational computer vision task that involves identifying multiple semantic attributes of a person from an image, such as gender, clothing, and carried items. As a core technology, it provides essential support for downstream applications, including person retrieval in video surveillance~\cite{wang2021pedestrian}, auxiliary matching for person re-identification~\cite{rethinkingofpar}, and intelligent retail analytics~\cite{deepmar}. Consequently, developing efficient and robust PAR algorithms is of paramount importance.

Propelled by deep learning, particularly Convolutional Neural Networks, PAR has achieved remarkable progress. The evolution of this field began with pioneering works that established foundational deep learning paradigms. For instance, early influential models like DeepMAR~\cite{deepmar} first demonstrated the effectiveness of using a single CNN to jointly learn multiple attributes, tackling it as a multi-label classification problem. Building upon this, subsequent methods such as HP-Net~\cite{HPNET} introduced multi-task learning frameworks integrated with attention mechanisms, which significantly improved performance by enabling the model to focus on relevant image regions for different attributes.

Following these foundations, current state-of-the-art (SOTA) methods have further pushed the performance boundaries by addressing more nuanced challenges. Models incorporating Graph Neural Networks (GNNs)~\cite{li2019visual} were developed to explicitly model the complex inter-dependencies between attributes (e.g., the correlation between \textit{dress} and \textit{female}). Other approaches have focused on refining feature representations through sophisticated attention modules and stronger backbones. These advanced techniques have achieved exceptional accuracy on established academic benchmarks like PA-100K~\cite{2017pa100k}, PETA~\cite{deng2014pedestrian}, and EventPAR~\cite{EventPAR}, proving their powerful feature learning and recognition capabilities within controlled environments.

\begin{figure*}
\centering
\includegraphics[width=0.8\linewidth]{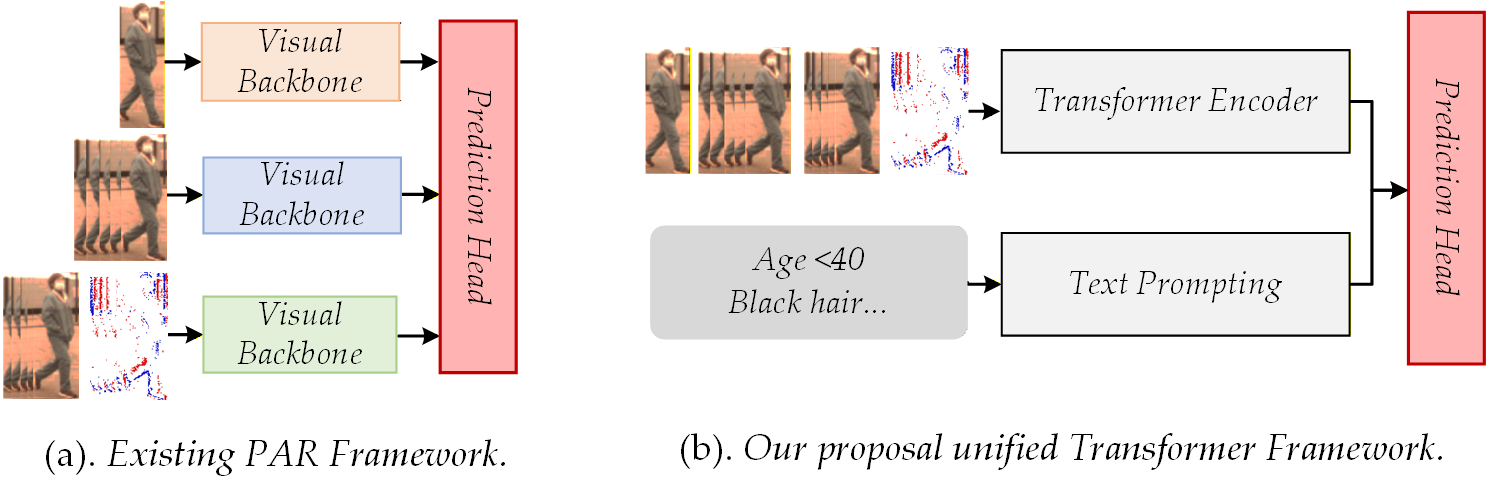}
\caption{Comparison between existing PAR models and our newly proposed one.}
\label{fig:comparison}
\end{figure*}

However, this benchmark success has not translated into real-world robustness. SOTA PAR models face significant challenges. The primary issue is a limited generalization capability; due to domain shift (e.g., variations in cameras, lighting), model performance drops sharply when deployed in unseen scenarios. This shortcoming also exposes the inefficiency of the prevailing ``one-model-per-dataset" paradigm, which is costly and difficult to scale. Furthermore, in the pursuit of peak performance on specific benchmarks, many models have become excessively complex and over-specialized. This practice of optimizing for benchmark scores ultimately sacrifices the flexibility and maintainability needed to adapt to new domains, further compounding the generalization problem.

To address these challenges and break the ``one-model-per-domain" bottleneck, we introduce a more universal paradigm for PAR. We propose a Unified Transformer Framework for Multi-Domain Pedestrian Attribute Recognition. Fig.~\ref{fig:comparison} highlights the architectural divergence between traditional PAR frameworks and the proposed UniPAR. Unlike the prevailing ``one-model-per-dataset" approach, our framework introduces a Phased Fusion Encoder to standardize attribute recognition across diverse modalities and datasets. The core idea is to leverage a single model, jointly trained on multiple heterogeneous datasets, to learn a more robust and generalizable visual-semantic alignment. Our main contributions are as follows:

\begin{enumerate}
    \item We introduce a \textbf{Novel, Unified, Transformer-based PAR Model} capable of end-to-end joint training across multiple datasets from diverse domains (e.g., RGB images and event-based data). At its core is an innovative Phased Fusion Encoder. This encoder first leverages deep Transformer layers for comprehensive visual understanding and then introduces textual attribute queries only at the final stage for deep cross-modal fusion. This ``late deep fusion" strategy ensures the model establishes a strong visual representation before using semantic cues to precisely locate corresponding visual evidence.

    \item We design a \textbf{Unified Data Scheduling Strategy} and a \textbf{Dynamic Classification Head}. The former ensures stability and efficiency during multi-dataset joint training via independent caching and sampling mechanisms. The latter enables the single model to flexibly adapt to the varying attribute categories and counts across different datasets, significantly enhancing the framework's scalability.

    \item We have conducted extensive experiments on several major benchmarks, including MSP60K, DukeMTMC-Attribute, and the emerging event-based dataset EventPAR, to validate our method's effectiveness. The results demonstrate that our unified model achieves performance that is not only comparable to specialized SOTA methods trained on individual datasets but also shows superior \textbf{Cross-Domain Generalization}. This highlights its significant potential as a universal solution for PAR.
\end{enumerate}

\section{Related Works} 

\subsection{Pedestrian Attribute Recognition}  
Pedestrian Attribute Recognition~\cite{wang2021pedestrian} aims to interpret the visual characteristics of pedestrians and stands as a key technology in fields like intelligent surveillance and human-computer interaction. The technological evolution in this domain clearly reflects the broader trend in computer vision, moving from specialized models toward general-purpose foundation models. Initially, researchers employed Convolutional Neural Networks (CNNs) to extract local features, often combined with Recurrent Neural Networks (RNNs) to capture sequential relationships between attributes~\cite{Zhang_2016}. However, these methods were limited in their global modeling capabilities. The introduction of the Transformer architecture fundamentally changed this landscape~\cite{VIT}. Its self-attention mechanism could effectively correlate global information across an image, and models like VTB~\cite{VTB} significantly improved recognition accuracy and robustness by fusing visual and textual modalities. 

Subsequently, the advent of large-scale Vision-Language Models (VLMs)~\cite{bordes2024introductionvisionlanguagemodeling}~\cite{li2022blip}~\cite{zhai2023sigmoidlosslanguageimage} propelled PAR research to new heights. Instead of training models from scratch, researchers began exploring how to efficiently transfer the general knowledge of VLMs to the PAR task. PromptPAR~\cite{PromptPAR} achieved remarkable performance by freezing most model parameters and using an innovative region-aware prompt tuning technique, validating the immense potential of the ``pre-train, fine-tune" paradigm. Building on this, LLM-PAR~\cite{LLMPAR} integrated the reasoning abilities of Large Language Models (LLMs) into the framework, enhancing the model's semantic understanding through natural language descriptions. This marked the beginning of PAR models acquiring higher-level cognitive abilities. In pursuit of higher performance, the efficiency and applicability of the model have also become focal points. Research has demonstrated that State Space Models (SSMs) like Mamba can significantly reduce computational complexity while maintaining high performance~\cite{wang2024mambayolo}~\cite{hatamizadeh2025mambavision}~\cite{wang2024MambaPAR}. On the application front, studies are expanding from static images to more complex, real-world scenarios, exemplified by VTFPAR++~\cite{VTFPAR++} for video data and EventPAR~\cite{EventPAR}, which fuses event streams to handle extreme lighting conditions. Currently, initial investigations into model security have provided a warning for the construction of reliable PAR systems~\cite{kong2025ASLPAR}~\cite{chen2025pass_icml}.

Observing this developmental trajectory—from the feature engineering of CNNs or RNNs, to the global modeling of Transformers, and onto the knowledge-driven approaches of VLMs and LLMs—it is clear that PAR technology is gradually shifting from solving singular, specific tasks towards the construction of a general, adaptable, and unified model. The future PAR system may no longer be a specialized network trained for a specific dataset or attribute list. Instead, it could be a unified foundation model capable of processing multi-modal inputs (e.g., images, videos, event streams), understanding task requirements via natural language prompts (e.g., identifying standard attributes, emotions, or interaction intents), and delivering structured analysis in an efficient and secure manner~\cite{qin2025facehumanbenchcomprehensivebenchmarkface}~\cite{wu2025enhanced}. This trend compels us to focus not only on innovating model architectures but also on how to achieve the diversification and synergy of data, tasks, and objectives within a unified framework, thereby taking a firm step toward general artificial intelligence.

\subsection{Unified Modelling}
In recent years, the research focus in Pedestrian Attribute Recognition has shifted from optimizing task-specific models on individual benchmarks towards developing unified frameworks with stronger generalization capabilities. This transition was initially driven by unification at the dataset level. To address the poor generalization of models trained on isolated datasets such as PA100K~\cite{2017pa100k} and PETA~\cite{deng2014pedestrian}, the community has constructed unified or cross-domain benchmarks like MSP60K~\cite{LLMPAR}, which introduce synthetic degradations to simulate real-world challenges. This has propelled a focus on domain generalization and led to the emergence of frameworks like AdaGPAR~\cite{li2025adagpar}, which leverage Test-Time Adaptation (TTA) strategies to adjust models online.

At the model architecture level, the trend towards unification evolved from early multi-task learning approaches, such as jointly training PAR with person re-identification (Re-ID), to building large-scale, general-purpose ``Human-Centric Perception" foundation models. For instance, works like UniHCP~\cite{ci2023unihcp} and HumanBench~\cite{tang2023humanbench} utilize a unified Transformer architecture to simultaneously handle multiple tasks including PAR, through task-specific queries or large-scale joint pre-training on as many as 37 datasets, 
aiming to learn generalizable cross-task human representations.
The unification of architectures and modalities has also become a mainstream trend, with the research paradigm shifting from customized networks to foundation models like the Vision Transformer. New paradigms have also emerged, such as SequencePAR~\cite{jin2023sequencePAR}, which recasts PAR as a sequence generation task. Furthermore, large-scale vision-language models  have opened new avenues for enhancing generalization. Frameworks like AttriVision~\cite{AttriVision} leverage the powerful vision-language alignment capabilities of CLIP, while advanced prompt learning-based methods such as PromptPAR~\cite{PromptPAR}, EVSITP~\cite{wu2025evsitp}, and ViTA-PAR~\cite{park2025vitapar} enhance modal interaction and fine-grained feature capture through dynamic, region-aware, or multi-modal prompts. Concurrently, frameworks like LLM-PAR~\cite{LLMPAR} directly utilize Large Language Models (LLMs) for ensemble learning to boost recognition performance. To fundamentally overcome the physical limitations of traditional RGB sensors, the latest research has begun to explore the unification of sensor modalities. The release of the EventPAR~\cite{EventPAR} dataset marks a milestone, introducing the first large-scale RGB-Event dual-modal data and extending attribute definitions to include the dimension of human emotions. The current unified models, while powerful, struggle with a trade-off between broad task generality and fine-grained feature specificity, especially in nuanced tasks like PAR, leading to potential inter-task interference. Additionally, existing visual-language models and RGB-Event sensor fusion methods lack deep, dynamically adaptive cross-modal interactions, limiting their efficiency and real-world applicability.


\begin{figure*}
\centering
\includegraphics[width=1\linewidth]{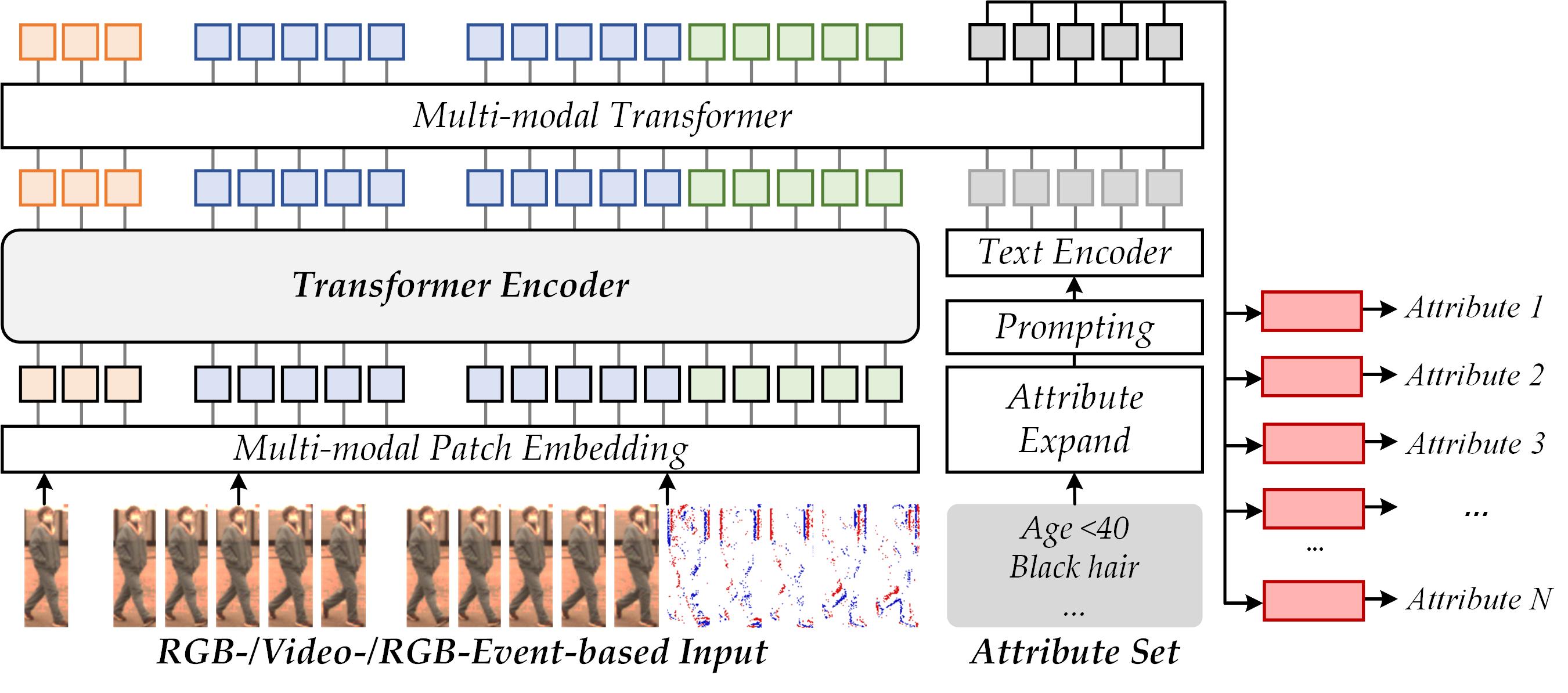}
\caption{The overall architecture of our proposed UnifiedPAR framework}
\label{fig:UnifiedPAR_framework}
\end{figure*}

\section{Methodology} 

\subsection{Overview} 
To address the challenges in pedestrian attribute recognition, namely heterogeneous datasets, modality diversity, and attribute imbalance, we design and implement a Unified Multi-modal Multi-dataset framework for Pedestrian Attribute Recognition. The core concept of this framework is to construct a single model capable of seamlessly processing data from various sources and modalities, such as RGB images, video sequences, and event streams. This is achieved through a unique data scheduling strategy and a weighted loss mechanism to facilitate efficient and stable joint training. Our proposed UniPAR architecture is based on the Vision Transformer but is deeply customized to achieve a late and deep fusion of visual features and textual attribute queries. This design enables the model not only to extract powerful visual representations but also to explicitly align visual evidence with specific attribute semantics. The overall network architecture is depicted in Figure~\ref{fig:UnifiedPAR_framework}, and its forward propagation process consists of three primary stages.

\subsection{Multi-modal Visual Embedding} 

To achieve unified processing of heterogeneous inputs, the network's entry point is a highly customized multi-modal embedding module. This process begins with modality-specific ``stems'', which are independent 2D convolutional layers configured for different modalities like RGB and event streams to perform initial patch embedding. This ensures that the model learns unique low-level patterns from each data type. After patch embedding, these token sequences are enriched with spatio-temporal and modality information. Specifically, all visual tokens are augmented with a learnable spatial position embedding, $E_{\text{spat}}$, to encode their relative positions in a 2D plane. For video or event sequences, a temporal position embedding, $E_{\text{temp}}$, is additionally incorporated to capture the sequential order of the frames. To maintain modality differentiability within the unified feature space, tokens from auxiliary modalities, such as event streams, are marked with a unique modality type embedding, $E_{\text{mod}}$. Furthermore, to efficiently handle temporal inputs with multiple frames, we introduce a lightweight Time Adapter. This adapter, composed of a multi-layer perceptron, can fuse and compress features from multiple frame tokens, significantly improving computational efficiency while preserving key dynamic information. Through this series of operations, any form of visual input is standardized into a visual token sequence rich in information, $F_{\text{vis}_0}$. 
\begin{equation}
 F_{\text{vis}_0} = \text{PatchEmbed}(X_{\text{vis}}) + E_{\text{spat}} + E_{\text{temp}} + E_{\text{mod}} 
\end{equation}

\noindent $\bullet$ \textbf{Phased Fusion Encoder}
The core of our framework's design lies in the phased fusion encoder, which facilitates the effective alignment of visual information and textual queries. We do not simply feed the visual tokens into a complete Transformer encoder. Instead, we creatively divide a pre-trained ViT backbone of $L$ layers into two functionally distinct stages. In the first stage, the visual token sequence $F_{\text{vis}_0}$ passes through the first $L-1$ Transformer encoder layers. During this phase, the model focuses on modeling deep visual context, capturing complex global and local relationships within the image or video through multi-layer self-attention. This process yields a comprehensive and unbiased understanding of the scene and target, resulting in refined visual features, $F_{\text{vis}}$.
\begin{equation}
 F_{\text{vis}} = \text{Encoder}_{1:L-1}(F_{\text{vis}_0})
\end{equation}

Subsequently, in the second stage, we introduce textual query tokens, $T_{\text{attr}}$, which represent the attribute semantics (composed of attribute word vectors). These tokens are concatenated with the visual features $F_{\text{vis}}$ from the first stage along the sequence dimension. We deliberately employ this late fusion design, as its fundamental advantage is allowing the model to first fully comprehend ``what is seen" before addressing ``what to look for" based on the textual instructions. This concatenated fusion sequence is then fed into the final encoder layer, $\text{Encoder}_{L}$. At this terminal layer, the self-attention mechanism performs the critical task of cross-modal alignment. The textual attribute tokens act as ``queries" to actively attend to relevant regions within the visual feature token sequence, and the visual tokens respond differently to various attribute queries. This interaction dynamically ``locates" the most relevant visual evidence for each attribute within the image.
\begin{equation}
 F_{\text{fused}} = \text{Encoder}_{L}([F_{\text{vis}} ; T_{\text{attr}}]) 
\end{equation}

\noindent $\bullet$ \textbf{Dynamic Classification Head}
The final classification task is accomplished by a dynamic classification head. To avoid designing a complex, unified output layer for datasets with varying numbers of attributes, we predefine a set of independent linear classification layers and Batch Normalization layers for each dataset. Each set is perfectly matched to the number of attributes in its corresponding dataset. During the model's forward pass, it dynamically routes the output to the appropriate classification layers based on the dimension of the input textual query tokens, $T_{\text{attr}}$ (i.e., the number of attributes). This design not only maintains the framework's simplicity but also ensures that the classification task for each dataset is performed within its own dedicated, optimal parameter space, thereby maximizing the effectiveness of joint training. For the fused feature token $f_j$ corresponding to the attribute $j$-th, its final prediction probability $p_j$ is calculated as follows:
\begin{equation}
 p_j = \sigma(\text{BN}(\mathbf{W}_j f_j)) 
\end{equation}
where $\mathbf{W}_j$ is the weight matrix of the linear layer associated with the attribute $j$-th, BN is the Batch Normalization operation, and $\sigma$ is the Sigmoid activation function.

\subsection{Unified Data Scheduling Strategy}

To enable the unified architecture to learn efficiently from multiple heterogeneous datasets, we have designed a synergistic data scheduling strategy that encompasses both training and validation phases.

During the training phase, we introduce a novel ``divert-cache-train-on-demand" mechanism to resolve the training instability that can arise from directly mixing data from different sources. This process begins with a ``universal data adapter", a custom collation function responsible for standardizing samples of various formats from different datasets (e.g., padding labels of different lengths, un-nesting multi-frame samples into a flat list of images) into a uniform tensor format. Each sample is also tagged with its source dataset ID. The adapted mixed-batch data is not fed directly into the model. Instead, samples are diverted to their corresponding First-In-First-Out (FIFO) cache queues based on their source ID. An independent training engine operates in an opportunistic manner, continuously monitoring the status of all cache queues. As soon as any queue accumulates enough samples to form a complete, single-source batch, that batch is immediately retrieved and used for one full training iteration (forward, backward, and parameter update). This asynchronous, on-demand training method ensures that the gradients received by the model always originate from a batch with a pure data distribution and consistent statistical properties, thereby greatly enhancing the stability and efficiency of the multi-task joint training.

In contrast to the mixed strategy of the training phase, our goal during the validation phase is to obtain clear and independent performance metrics for each dataset. To this end, we employ a rigorous rotational evaluation mechanism. After each training epoch, the evaluation procedure sequentially iterates through all configured datasets. When a specific data set is being evaluated, the validation data loader is explicitly configured to provide only the validation samples of that particular data set. Upon completion of the evaluation for one dataset, the process switches to the next until all datasets have been accurately and independently evaluated.
\subsection{Objective Function}

Pedestrian attribute recognition, as a typical multi-label classification task, commonly suffers from class imbalance. To address this issue and to allow the objective function to adapt to the unique attribute distributions of different datasets, we employ a dataset-aware weighted binary cross-entropy loss. For a batch of $N$ samples, the total loss $\mathcal{L}$ is defined as the average of the weighted losses over all samples and all attributes:
\begin{equation}
\mathcal{L} = \frac{1}{N} \sum_{i=1}^{N} \sum_{j=1}^{C} w_j \cdot \ell_{\text{BCE}}(p_{ij}, y_{ij}) 
\end{equation}
where $C$ is the number of attributes for the dataset to which the current batch belongs, $y_{ij} \in \{0, 1\}$ and $p_{ij}$ are the ground-truth label and predicted probability for the $j$-th attribute of the $i$-th sample, respectively, and $\ell_{\text{BCE}}$ is the standard binary cross-entropy loss. The core of this function is the dataset-specific weight vector, $\mathbf{w} = \{w_1, w_2, \dots, w_C\}$. This vector is computed independently for each dataset. The weight $w_j$ is inversely proportional to the occurrence frequency (positive rate) $r_j$ of its corresponding attribute in the training set, determined, for example, by a smoothed inverse function such as $w_j = \log(1/r_j + 1)$. This design, where each dataset has its own system of weights, is crucial. It ensures that an attribute, even if its distribution frequency differs vastly across datasets, receives appropriate optimization focus within its own context, thus achieving fine-grained, adaptive optimization within the unified framework.

\section{Experiments} 

\subsection{Datasets and Evaluation Metric}  

To comprehensively evaluate the performance of our proposed pedestrian attribute recognition method, we conducted experiments on three widely-used public datasets, including \textbf{MSP60K}~\cite{LLMPAR}, \textbf{DukeMTMC-VID-Attribute}~\cite{duke}, and \textbf{EventPAR}~\cite{EventPAR}.

$\bullet$ \textbf{MSP60K:} This is a large-scale, cross-domain benchmark dataset for Pedestrian Attribute Recognition, introduced by Wang et al. in 2024. The dataset comprises 60,122 images covering eight different scenarios and includes detailed annotations for 57 attributes. A key contribution of this dataset is the introduction of synthetically degraded images to actively bridge the gap between the dataset and complex real-world visual conditions.

$\bullet$ \textbf{DukeMTMC-VID-Attribute:} This dataset originates from a multi-camera surveillance network widely used for the Person Re-identification (Re-ID) task. For our attribute recognition task, we use its version with attribute annotations. The dataset includes over 4,000 samples and provides 36 distinct attribute labels, making it highly suitable for evaluating model performance in complex surveillance environments.

$\bullet$ \textbf{EventPAR:} To test the model's robustness on data from novel sensors, we also utilize the EventPAR dataset. It is specifically designed for pedestrian data collected by event cameras and contains over 10,000 samples. The dataset provides a rich set of 50 attribute annotations, which is crucial for validating algorithm performance in unconventional scenarios, such as those with high dynamic range and low-light conditions.

To evaluate the performance of our method, we adopt three standard metrics: \textbf{mean Accuracy (mA)}, \textbf{Accuracy (Acc)}, and \textbf{F1-score (F1)}. mA is particularly important as it calculates the average accuracy for each attribute, offering an unbiased evaluation on imbalanced datasets.

\subsection{Implementation Details}  
The proposed framework is implemented using the PyTorch~\cite{paszke2019pytorch} deep learning library. To ensure a robust representation for pedestrian attribute recognition, all input images—including both static RGB images and event-stream frames—are uniformly resized to a resolution of $256 \times 128$ pixels. During the training phase, we employ a comprehensive data augmentation suite to mitigate overfitting, including random horizontal flipping, padding, random cropping, and random erasing. For datasets involving event data, such as EventPAR and DukeMTMC-VID-Attribute, the temporal stream is processed via a stack of five synchronized frames to capture motion-related cues.

The optimization process is driven by the AdamW optimizer with a weight decay of $1 \times 10^{-4}$. We adopt a learning rate scheduling strategy that combines a linear warm-up (initial 5 epochs) with a cosine annealing decay. The base learning rate is set to $8 \times 10^{-3}$, and the total training duration is fixed at 60 epochs. To address the challenge of joint training across heterogeneous datasets, we utilize a mixed-batch sampling strategy and a custom collation mechanism. This allows the model to simultaneously handle diverse data modalities—ranging from single RGB images to multi-modal video sequences—within a unified batch. Furthermore, to accommodate varying attribute spaces across different datasets, we implement a masked classification loss with sigmoid activation, where loss weights are adaptively adjusted based on the sample distribution of each attribute.

All experiments are conducted on an NVIDIA RTX 4090D GPU with a batch size of 8 per dataset (effectively doubled during mixed-batch training). For the feature extractor, we utilize a Vision Transformer architecture with a hidden dimension of 768. To ensure the reproducibility of our results, we set a fixed seed of 605 for all stochastic operations. More details can be found in our source code.

\begin{table*}[h]
\centering
\caption{The experimental result of our proposed method.} 
\label{tab:results}
\begin{tabular}{c|l|ccccccc}
\toprule
&\textbf{Dataset}   &\textbf{mA} &\textbf{Accuracy} &\textbf{Precision} &\textbf{Recall} &\textbf{F1} \\
\midrule 
\multirow{3}{*}{\makecell{Individually \\ Trained}}  
& MSP60K~\cite{LLMPAR} & 75.12 & 76.32 & 86.77 & 84.32 & 85.15 \\
& DUKE~\cite{duke} & 69.73 & 61.76 & 75.58 & 73.78 & 74.09 \\
& EventPAR~\cite{EventPAR}  & 86.90 & 82.31 & 87.01 & 88.41 & 87.53 \\
\bottomrule
\multirow{3}{*}{\makecell{Jointly \\ Trained}}  
& MSP60K~\cite{LLMPAR}      & 79.55 & 77.79 & 85.86 & 87.54 & 86.32  \\
& DUKE~\cite{duke}          & 75.56	& 70.40	& 82.77 & 79.92 & 80.75  \\ 
& EventPAR~\cite{EventPAR}  & 88.51	& 85.21 & 89.22 & 89.77 & 89.36  \\ 
\bottomrule
\end{tabular}
\end{table*}

\begin{table*}[t]
\center
\setlength{\tabcolsep}{5pt}
\small
\caption{Comparison with state-of-the-art methods on MSP60K datasets.}
\label{tab:comparisonMSP60k}
\begin{tabular}{l|l|c|ccccc}
\hline \toprule [0.5 pt]
\multirow{1}{*}{\textbf{Methods}} & 
\multirow{1}{*}{\textbf{Publish}} & 
\multirow{1}{*}{\textbf{Code}} & 
\textbf{mA} & \textbf{Acc} & \textbf{Prec} & \textbf{Recall} & \textbf{F1} \\ 
\hline
\#01 DeepMAR~\cite{deepmar} & ACPR15 &  \href{https://github.com/dangweili/pedestrian-attribute-recognition-pytorch}{URL} & 70.46 & 72.83 & 84.71 & 81.46 & 83.06 \\
\#02 Strong Baseline~\cite{rethinkingofpar} & arXiv20 & \href{https://github.com/aajinjin/Strong_Baseline_of_Pedestrian_Attribute_Recognition}{URL} & 74.09 & 73.74 & 84.06 & 83.51 & 83.31 \\
\#03 RethinkingPAR~\cite{rethinkingofpar} & arXiv20 & \href{https://github.com/valencebond/Rethinking_of_PAR}{URL} & 74.01 & 74.20 & 84.17 & 83.94 & 84.06 \\
\#04 SSCNet~\cite{jia2021spatial} & ICCV21 & \href{https://github.com/Geonu-Lee/Reimplementation_SSC}{URL} & 69.71 & 69.31 & 79.22 & 82.47 & 80.82 \\
\#05 VTB~\cite{VTB}  & TCSVT22  & \href{https://github.com/cxh0519/VTB}{URL} & 76.09 & 75.36 & 83.56 & 86.46 & 84.56 \\
\#06 Label2Label~\cite{label2label} & ECCV22 & \href{https://github.com/Li-Wanhua/Label2Label}{URL} & 73.61 & 72.66 & 81.79 & 84.32 & 82.56 \\
\#07 DFDT~\cite{zheng2023diverse} & EAAI22 & \href{https://github.com/aihuazheng/DFDT}{URL} & 74.19 & 76.35 & 85.03 & 86.35 & 85.69 \\
\#08 Zhou et al.~\cite{zhou2023} & IJCAI23 & \href{https://github.com/SDret/A-Solution-to-Co-occurrence-Bias-Attributes-Disentanglement-via-Mutual-Information-Minimization-for}{URL} & 73.07 & 68.76 & 78.38 & 82.10 & 80.20 \\
\#09 PARFormer~\cite{fan2023parformer} & TCSVT23 & \href{https://github.com/xwf199/PARFormer}{URL} & 76.14 & 76.67 & 84.77 & 86.93 & 85.44 \\
\#10 SequencePAR~\cite{jin2023sequencePAR} & arXiv23   &  \href{https://github.com/Event-AHU/OpenPAR}{URL} & 71.88 & 71.99 & 83.24 & 82.29 & 82.29 \\
\#11 VTB-PLIP ~\cite{zuo2024plip} & arXiv23  & \href{https://github.com/Zplusdragon/PLIP}{URL} & 73.90 & 73.16 & 82.01 & 84.82 & 82.93 \\
\#12 Rethink-PLIP~\cite{zuo2024plip}  & arXiv23  & \href{https://github.com/Zplusdragon/PLIP}{URL} & 69.44 & 68.90 & 79.82 & 81.15 & 80.48 \\
\#13 PromptPAR~\cite{wang2024promptpar}  &TCSVT24  &  \href{https://github.com/Event-AHU/OpenPAR}{URL} & 78.81 & 76.53 & 84.40 & 87.15 & 85.35 \\
\#14 SSPNet~\cite{sspnet} & PR24 &  \href{ https://github.com/guotengg/SSPNet}{URL} & 74.03 & 74.10 & 84.01 & 84.02 & 84.02 \\
\#15 HAP~\cite{HAP}  & NIPS24  & \href{https://github.com/junkunyuan/HAP}{URL} & 76.92 & 76.12 & 84.78 & 86.14 & 85.45 \\
\#16 MambaPAR~\cite{wang2024MambaPAR} & arXiv24 & \href{https://github.com/Event-AHU/OpenPAR/tree/main/MambaPAR_Empirical_Study}{URL} & 73.85 & 73.64 & 83.19 & 84.29 & 83.28 \\
\#17 MaHDFT~\cite{MaHDFT} & arXiv24  & \href{https://github.com/Event-AHU/OpenPAR/tree/main/MambaPAR_Empirical_Study}{URL} & 74.08 & 74.40 & 82.82 & 86.41 & 83.93 \\
\#18 LLM-PAR & AAAI25 & \href{https://github.com/Event-AHU/OpenPAR/tree/main/MSP60K_Benchmark_Dataset}{URL} & 80.13 & 78.71 & 84.39 & 90.52 & 86.94 \\
\hline
\rowcolor{red!10}
\textbf{Ours} & - & - &75.12 & 76.32 & 86.77 & 84.32 & 85.15 \\
\hline \toprule [0.5 pt]
\end{tabular}
\end{table*}

\begin{table*}[!htb]
\center
\setlength{\tabcolsep}{5pt}
\small  
\caption{Comparison with state-of-the-art methods on EventPAR dataset. }   
\label{tab:comparisonEevntPAR} 
\begin{tabular}{l|l|c|ccccc}
\hline \toprule [0.5 pt] 
\multirow{1}{*}{\textbf{Methods}} & 
\multirow{1}{*}{\textbf{Publish}} & 
\multirow{1}{*}{\textbf{Code}} &
\multicolumn{1}{c}{\textbf{mA}} &
\multicolumn{1}{c}{\textbf{Acc}} & 
\multicolumn{1}{c}{\textbf{Prec}} &
\multicolumn{1}{c}{\textbf{Recall}} &
\multicolumn{1}{c}{\textbf{F1}}
\\ \hline 
\#01~~DeepMAR~\cite{deepmar}  & ACPR15 & \href{https://github.com/dangweili/pedestrian-attribute-recognition-pytorch}{URL} &66.57  &69.53  &74.90  & 89.54   &81.57   \\
\#02~~ALM ~\cite{tang2019improving} & ICCV19  & \href{https://github.com/chufengt/ALM-pedestrian-attribute}{URL}  & 57.18 & 64.17 &75.59  & 73.20 &74.38    \\
\#03~~Strong Baseline~\cite{rethinkingofpar} &arXiv20 & \href{https://github.com/aajinjin/Strong_Baseline_of_Pedestrian_Attribute_Recognition}{URL}  & 73.75 &61.86  &67.23  &80.78  &75.43  \\
\#04~~RethinkingPAR~\cite{rethinkingofpar} & arXiv20 & \href{https://github.com/valencebond/Rethinking_of_PAR}{URL} & 81.37 & 80.84 & 86.31 & 87.57 & 86.93 \\
\#05~~SSCNet~\cite{jia2021spatial} & ICCV21 & \href{https://github.com/Geonu-Lee/Reimplementation_SSC}{URL}& 63.10 &66.07  &72.72  & 83.22 & 77.62     \\
\#06~~VTB~\cite{VTB}  & TCSVT22  & \href{https://github.com/cxh0519/VTB}{URL} & 88.41 & 83.83 &87.89  &89.51  & 88.53   \\
\#07~~Label2Label~\cite{label2label} & ECCV22 & \href{https://github.com/Li-Wanhua/Label2Label}{URL} &72.49  &74.01  &86.60  &79.02  &82.19      \\
\#08~~DFDT~\cite{zheng2023diverse} & EAAI22 & \href{https://github.com/aihuazheng/DFDT}{URL} & 61.71 & 63.14 & 79.17 & 7.63 & 74.66    \\
\#09~~Zhou et al.~\cite{zhou2023} & IJCAI23 & \href{https://github.com/SDret/A-Solution-to-Co-occurrence-Bias-Attributes-Disentanglement-via-Mutual-Information-Minimization-for}{URL} 
& 56.46 & 60.89 &73.37  & 73.62 & 73.50 \\
\#10~~PARFormer~\cite{fan2023parformer} & TCSVT23 & \href{https://github.com/xwf199/PARFormer}{URL}  &83.12  & 80.48 &85.14  &  88.41&86.53    \\ 
\#11~~SequencePAR~\cite{jin2023sequencePAR} & arXiv23   &  \href{https://github.com/Event-AHU/OpenPAR}{URL} & 86.27 &84.42  & 88.81 &89.12  &88.83   \\
\#12~~VTB-PLIP ~\cite{zuo2024plip} & arXiv23  & \href{https://github.com/Zplusdragon/PLIP}{URL}  &67.25  &68.37  & 77.75 &79.72  &78.37    \\
\#13~~Rethink-PLIP~\cite{zuo2024plip}  & arXiv23  & \href{https://github.com/Zplusdragon/PLIP}{URL}  & 68.75 & 70.03 & 81.82 & 78.04 & 79.89   \\
\#14~~PromptPAR~\cite{wang2024promptpar}  &TCSVT24  &  \href{https://github.com/Event-AHU/OpenPAR}{URL} & 86.51 &82.27  & 86.35 &89.36  & 87.64     \\
\#15~~SSPNet ~\cite{sspnet} & PR24 &  \href{ https://github.com/guotengg/SSPNet}{URL} & 66.92 & 67.49 & 78.73 & 76.90 & 77.80   \\
\#16~~HAP~\cite{HAP}  & NIPS24  & \href{https://github.com/junkunyuan/HAP}{URL} &88.28  &85.42  & 89.54 & 89.72 & 89.63  \\
\#17~~MambaPAR~\cite{wang2024MambaPAR} & arXiv24 & \href{https://github.com/Event-AHU/OpenPAR/tree/main/MambaPAR_Empirical_Study}{URL} & 50.01 & 42.32 & 54.81 & 57.31 & 55.63   \\ 
\#18~~MaHDFT~\cite{MaHDFT} & arXiv24  & \href{https://github.com/Event-AHU/OpenPAR/tree/main/MambaPAR_Empirical_Study}{URL}  & 50.43 & 44.98 & 59.10 & 59.70 & 58.57    \\
\#19~~RWKV-OTN~\cite{EventPAR} &arXiv25 &\href{https://github.com/Event-AHU/OpenPAR}{URL} &87.70&84.94 &89.15  &89.48 &89.18     \\ 
\hline
\rowcolor{red!10}
\textbf{Ours}& - & - &86.90 & 82.31 & 87.01 & 88.41 & 87.53    \\ 
\hline \toprule [0.5 pt] 
\end{tabular}
\end{table*}

\subsection{Comparison on Public Benchmarks} 

To verify the effectiveness and generalization of the proposed UniPAR, we report its performance under both individual and joint training settings and compare it with state-of-the-art methods on two major benchmarks: MSP60K and EventPAR. As shown in Table~\ref{tab:results}, the joint training paradigm significantly boosts performance across all three datasets compared to individual training. Specifically, on the MSP60K dataset, the mean Accuracy (mA) increases from 75.12\% to 79.55\%, and the F1-score improves from 85.15\% to 86.32\%. Similar upward trends are observed in the DUKE and EventPAR datasets. These results demonstrate that our unified framework effectively leverages heterogeneous data to learn more robust and transferable visual-semantic representations, effectively breaking the ``one-model-per-dataset" bottleneck.

Table~\ref{tab:comparisonMSP60k} presents the comparison with representative SOTA methods on the MSP60K benchmark. Our UniPAR achieves competitive results with an mA of 75.12\% and an F1-score of 85.15\%. Notably, our model outperforms classic CNN-based methods like DeepMAR (70.46\% mA) and recent Transformer-based models such as PARFormer (76.14\% mA) and SequencePAR (71.88\% mA) in terms of overall accuracy and F1 metrics. While the Large Language Model augmented framework shows a higher mA (80.13\%), UniPAR maintains a superior balance between efficiency and accuracy without relying on heavy external LLM reasoning during inference.

Performance in the emerging EventPAR dataset is summarized in Table~\ref{tab:comparisonEevntPAR}. UniPAR achieves an impressive mA of 86.90\% and an F1-score of 87.53\%, significantly outperforming the baseline RWKV-OTN in specific precision metrics. Compared to methods like MambaPAR (50.01\% mA) and MaHDFT (50.43\% mA), which struggle with event-based data distributions, our model demonstrates superior robustness. The ``late deep fusion" strategy in our Phased Fusion Encoder allows the model to precisely locate visual evidence within high-dynamic-range event streams using semantic queries.

\subsection{Ablation Study} 
To evaluate the contribution of each key component in UniPAR, we conduct comprehensive ablation studies. Specifically, we investigate: (1) the effectiveness of the data unification strategy through cross-dataset joint training, and (2) the impact of the attribute-set-based text encoder on capturing semantic correlations.

\noindent $\bullet$ \textbf{Ablation Study on the Data Unification Strategy.} 
We first verify the scalability of our framework by performing joint training on different dataset pairs. As reported in Table \ref{tab:resultTwodatasets}, when MSP60k and DUKE are trained jointly, the model achieves 80.06\% mA on MSP60k, which is superior to individual training results. Similarly, the combination of MSP60k and EventPAR yields high performance across both benchmarks (79.34\% and 88.78\% mA respectively). These results indicate that our unification strategy successfully mitigates the domain gap between traditional RGB datasets and other-based datasets, allowing the model to learn complementary features from diverse data sources.

\begin{table*}[t]
\centering 
\small
\setlength{\tabcolsep}{15pt} 
\caption{Results of Cross-dataset Evaluation on MSP60k, DUKE, and EventPAR.}
\label{tab:resultTwodatasets}
\begin{tabular}{l|ccccc}
\toprule
\textbf{Jointly Trained} & \textbf{mA} & \textbf{Accuracy} & \textbf{Precision} & \textbf{Recall} & \textbf{F1 score} \\
\midrule
MSP60k & 80.06 & 77.65 & 85.51 & 87.63 & 86.18 \\
DUKE & 73.06 & 68.18 & 81.04 & 78.79 & 79.24 \\
\midrule 
MSP60k & 79.34 & 77.01 & 84.90 & 87.52 & 85.80 \\
EventPAR & 88.78 & 85.20 & 89.42 & 89.55 & 89.35 \\
\midrule 
DUKE & 74.13 & 66.38 & 79.10 & 77.54 & 77.68 \\
EventPAR & 88.87 & 83.47 & 87.78 & 89.15 & 88.30 \\
\bottomrule
\end{tabular}
\end{table*}

\noindent $\bullet$ \textbf{Ablation Study on Text Encoder.}
To verify the necessity of semantic guidance and explore the impact of different text representation methods, we compare our complete model with several variants in Table \ref{tab:resultsAttSet}: (1) \textit{Without Attribute Set}, which removes all text-based semantic embeddings; (2) \textit{BERT Embedding}, which utilizes a frozen BERT-base model to extract attribute vectors; (3) \textit{CLIP Embedding}, which employs the pre-trained CLIP text encoder.

The results demonstrate that any form of semantic guidance significantly boosts performance compared to the baseline. Although using standardized BERT or CLIP embeddings provides substantial improvements, our Full Model, which adopts a dataset-specific optimized encoding strategy, achieves the best results across almost all metrics. Notably, on the DUKE dataset, the Full Model reaches an mA of 75.56\% and an F1 score of 80.75\%, outperforming the CLIP-based variant. This proves that tailoring the encoding mechanism to specific dataset characteristics allows for more precise alignment between visual features and high-level concepts, especially in complex scenarios where visual cues might be ambiguous.

\begin{table*}[h]
\centering
\caption{The Ablation Study Result on Text Encoder.}
\label{tab:resultsAttSet}
\begin{tabular}{c|l|ccccccc}
\toprule
&\textbf{Dataset}   &\textbf{mA} &\textbf{Accuracy} &\textbf{Precision} &\textbf{Recall} &\textbf{F1} \\
\midrule 
\multirow{3}{*}{\makecell{Without \\ Attribute Set}}  
& MSP60K~\cite{LLMPAR} & 77.33 & 71.51 & 80.09 & 84.64 & 81.84 \\
& DUKE~\cite{duke} & 69.04 & 59.26 & 75.71 & 71.13 & 72.26 \\
& EventPAR~\cite{EventPAR}  & 86.56 & 81.73 & 86.60 & 88.14 & 87.19 \\
\bottomrule
\multirow{3}{*}{\makecell{BERT\\ Embedding}}  
& MSP60K~\cite{LLMPAR} & 79.22 & 77.11 & 85.37 & 87.11 & 85.84 \\
& DUKE~\cite{duke} & 72.94 & 66.87 & 80.29 & 77.57 & 78.22 \\
& EventPAR~\cite{EventPAR}  & 88.47 & 85.05 & 89.20 & 89.61 & 89.26 \\
\bottomrule
\multirow{3}{*}{\makecell{CLIP \\ Embedding}}  
& MSP60K~\cite{LLMPAR} & 79.31 & 77.38 & 85.64 & 87.17 & 86.01 \\
& DUKE~\cite{duke} & 73.41 & 67.84 & 80.16 & 78.91 & 78.87 \\
& EventPAR~\cite{EventPAR}  & 88.82 & 85.31 & 89.35 & 89.78 & 89.43 \\
\bottomrule
\multirow{3}{*}{\makecell{Full Model}}  
& MSP60K~\cite{LLMPAR}      & 79.55 & 77.79 & 85.86 & 87.54 & 86.32  \\
& DUKE~\cite{duke}          & 75.56	& 70.40	& 82.77 & 79.92 & 80.75  \\ 
& EventPAR~\cite{EventPAR}  & 88.51	& 85.21 & 89.22 & 89.77 & 89.36  \\ 
\bottomrule
\end{tabular}
\end{table*}

\subsection{Parameter Analysis} 
To investigate the impact of different loss contribution scales during the joint training process, we conduct a parameter analysis by adjusting the loss weight ratios (denoted as $Lossrate$) for the three benchmarks: MSP60k, DUKE, and EventPAR. The experimental results are summarized in Table~\ref{tab:resultPR}.

As shown in the table, the model's performance is sensitive to the balance of gradients from different data sources. When using an equal weight ratio ($1:1:1$), the model achieves a balanced but sub-optimal performance. Reducing the weight of the first dataset ($0.8:1:1$) leads to a noticeable performance degradation, particularly on MSP60k. Interestingly, the configuration ($0.8:1:0.6$) yields the best overall results, with the mA for DUKE significantly improving to 75.56\% and the F1 score for MSP60k reaching 86.32\%. This suggests that appropriately balancing the loss weights helps the model focus on learning more robust features across heterogeneous domains. Consequently, we adopt ($0.8:1:0.6$) as the default setting.

\begin{table*}[t]
\centering 
\small
\setlength{\tabcolsep}{15pt} 
\caption{Results of Parameter Analysis.}
\label{tab:resultPR}
\begin{tabular}{ll|ccccc}
\toprule
\textbf{Lossrate} & \textbf{Dataset} & \textbf{mA} & \textbf{Accuracy} & \textbf{Precision} & \textbf{Recall} & \textbf{F1 score} \\
\midrule
\multirow{3}{*}{1,1,1} & MSP60k & 78.85 & 76.31 & 84.36 & 87.13 & 85.31 \\
& DUKE & 72.39 & 66.32 & 80.11 & 76.90 & 77.76 \\
& EventPAR & 88.24 & 85.08 & 89.20 & 89.63 & 89.28 \\
\midrule 
\multirow{3}{*}{0.8,1,1} & MSP60k & 76.75 & 73.54 & 81.94 & 85.88 & 83.40 \\
& DUKE & 71.02 & 64.53 & 78.01 & 76.46 & 76.46 \\
& EventPAR & 87.65 & 84.20 & 88.84 & 89.93 & 88.74 \\
\midrule 
\multirow{3}{*}{0.8,1,0.6} & MSP60k & 79.55 & 77.79 & 85.86 & 87.54 & 86.32 \\
& DUKE & 75.56 & 70.40 & 82.77 & 79.92 & 80.75 \\
& EventPAR & 88.51 & 85.21 & 89.22 & 89.77 & 89.36 \\
\bottomrule
\end{tabular}
\end{table*}

\subsection{Visualization} 

To intuitively demonstrate the effectiveness and robustness of UniPAR, we provide a visual analysis through two perspectives: multi-modal generalization and cross-dataset joint training synergy.

\begin{figure}
\centering
\includegraphics[width=\linewidth]{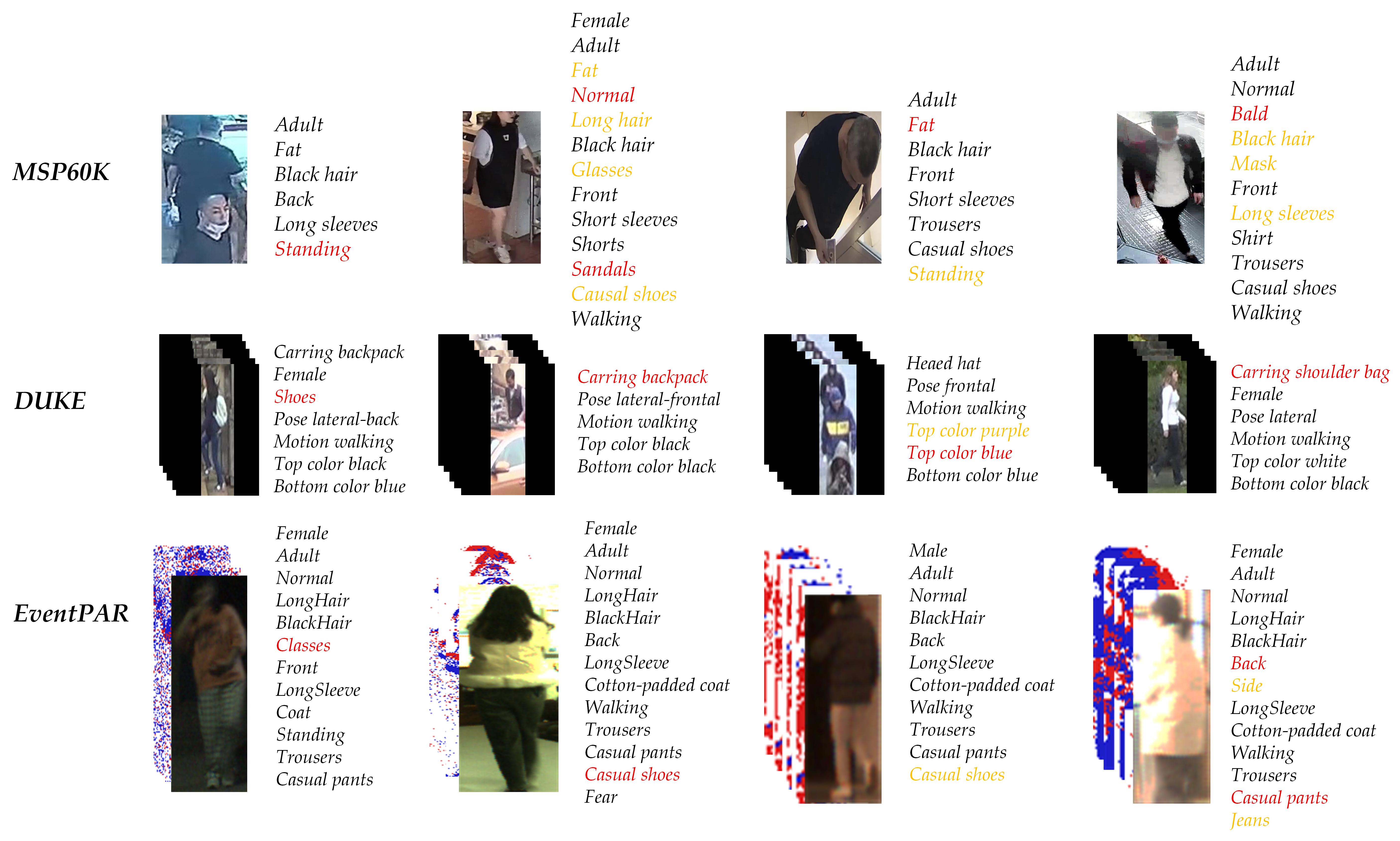}
\caption{Visual Analysis of the Generalization Performance of Our Proposed Model on the MSP60k, DUKE, and EventPAR Datasets. Red markers indicate misclassified attributes, while orange markers denote undetected attributes.}
\label{fig:UnifiedPAR_visual}
\end{figure}

As illustrated in Fig~\ref{fig:UnifiedPAR_visual}, we showcase the recognition results on the MSP60K, DUKE, and EventPAR datasets. The model exhibits robust performance across RGB images, video sequences, and event streams. In Fig~\ref{fig:UnifiedPAR_visual}, red markers indicate misclassified attributes, while orange markers denote undetected attributes. Despite the presence of minor errors in challenging samples—such as detecting ``Glasses" or ``Sandals" in low-light environments—the vast majority of semantic attributes are correctly identified. This proves that our Phased Fusion Encoder effectively leverages textual queries to anchor relevant visual evidence even when the input modality changes significantly.

\begin{figure*}[htbp]
\centering
\includegraphics[width=0.8\linewidth]{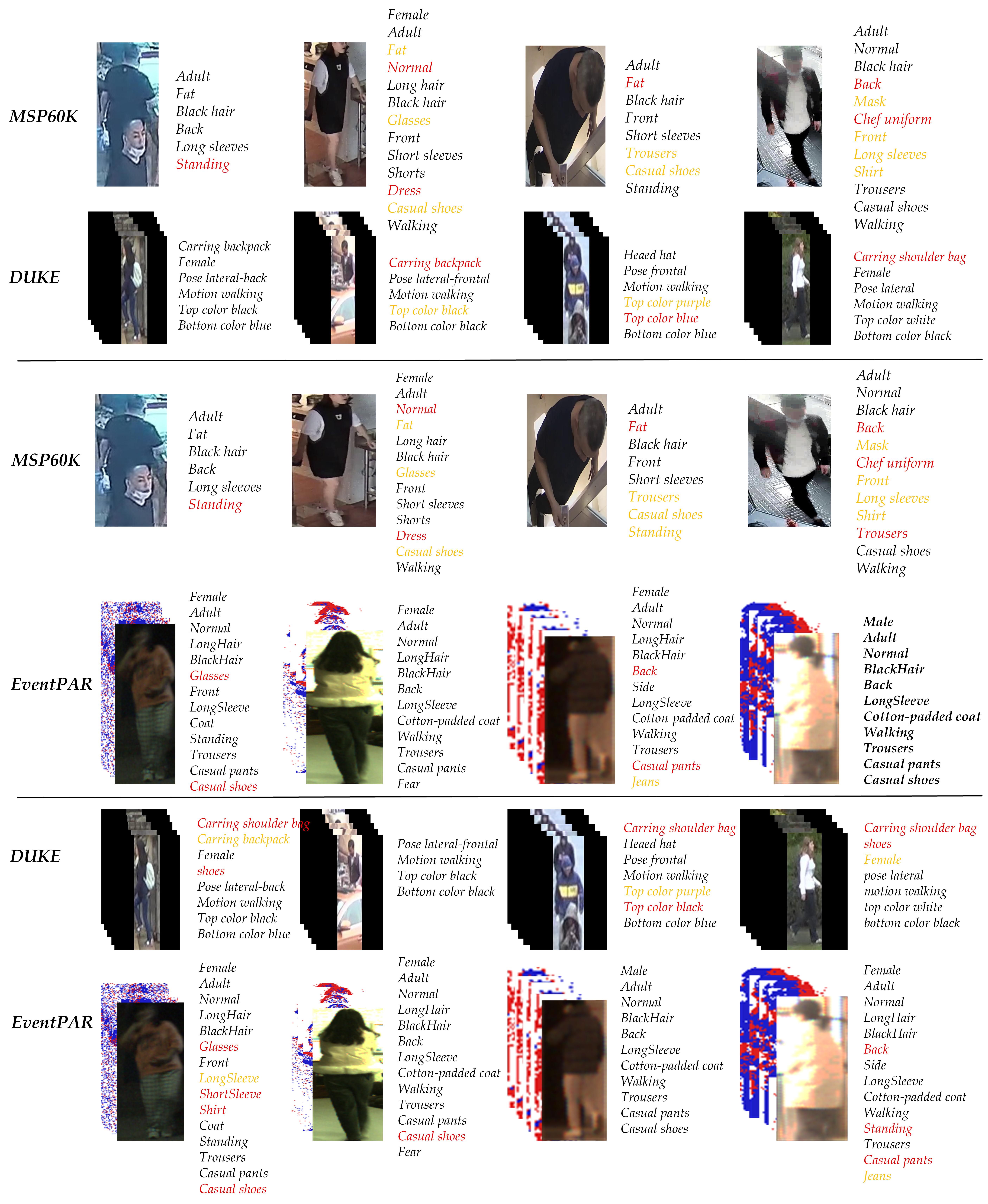}
\caption{Visualizing the Cross-Dataset Generalization of Our Proposed Model. Red markers indicate misclassified attributes, while orange markers denote undetected attributes.}
\label{fig:UnifiedPAR_twoDatasets}
\end{figure*}

Fig~\ref{fig:UnifiedPAR_twoDatasets} provides a qualitative comparison of the model's cross-dataset generalization capabilities. By visualizing the prediction results under joint training, we can observe how the model aligns visual features with high-level concepts across heterogeneous domains. The reduction of red and orange markers in the joint training setting (compared to individual training baselines) intuitively demonstrates that the ``one-model-for-all" paradigm mitigates domain shift. For instance, the model accurately captures ``Top color" and ``Action" across both DUKE and EventPAR samples. This visual evidence confirms that our unified framework successfully learns a more universal and transferable representation by diversifying the training data sources.
    
\subsection{Limitation Analysis}  

Despite the superior performance demonstrated by UniPAR across multiple benchmarks, a rigorous analysis identifies several avenues for further optimization. First, empirical results indicate that the model's performance on the EventPAR dataset significantly surpasses that on traditional RGB benchmarks. This discrepancy highlights the intrinsic advantages of event cameras in capturing dynamic pedestrian information, particularly in scenarios characterized by complex lighting or high-speed motion where standard RGB sensors inherently struggle. However, this also reveals a current limitation: the model's feature extraction capability in single-modality scenarios (e.g., pure RGB images in MSP60K) warrants further strengthening. At present, the model relies substantially on the synergistic effects derived from multi-dataset joint training to reach its peak performance. 

These findings suggest that while our unified framework successfully capitalizes on the benefits of event-based data, enhancing the backbone's ability to extract robust representations from individual modalities remains a primary objective. Furthermore, the success of Event-RGB fusion indicates that incorporating a broader range of modalities---such as infrared (IR) or depth information---could yield even more significant performance gains. In future research, we aim to optimize single-modality encoders to ensure consistently high recognition performance across all domains, even without multi-modal synergy. Additionally, transitioning the current dynamic classification head toward an open-vocabulary paradigm will further enhance the framework's scalability to unseen attribute categories.

\section{Conclusion}  

We proposed a unified Transformer-based framework designed to break the ``one-model-per-dataset" bottleneck prevalent in traditional Pedestrian Attribute Recognition. The model introduces an innovative Phased Fusion Encoder that explicitly aligns visual features with textual attribute queries through a ``late deep fusion" strategy. Furthermore, by incorporating a Unified Data Scheduling Strategy and a Dynamic Classification Head, UniPAR is capable of simultaneously processing heterogeneous datasets from diverse domains, containing various modalities (such as RGB, video, and event streams) and differing attribute definitions within a single model.

Experimental results demonstrate that UniPAR achieves performance comparable to specialized SOTA methods across mainstream benchmarks including MSP60K, DukeMTMC, and EventPAR. Notably, through multi-dataset joint training, the model exhibits exceptional cross-domain generalization, significantly enhancing recognition robustness in extreme environments such as low light and motion blur.

{
    \small
    \bibliographystyle{ieeenat_fullname}
    \bibliography{main}
}


\end{document}